\documentclass[10pt,a4paper,onecolumn]{article}
\usepackage{marginnote}
\usepackage{graphicx}
\usepackage{xcolor}
\usepackage{authblk,etoolbox}
\usepackage{titlesec}
\usepackage{calc}
\usepackage{tikz}
\usepackage{hyperref}
\hypersetup{colorlinks,breaklinks,
            urlcolor=[rgb]{0.0, 0.5, 1.0},
            linkcolor=[rgb]{0.0, 0.5, 1.0}}
\usepackage{caption}
\usepackage{tcolorbox}
\usepackage{amssymb,amsmath}
\usepackage{ifxetex,ifluatex}
\usepackage{seqsplit}
\usepackage{xstring}

\usepackage{float}
\let\origfigure\figure
\let\endorigfigure\endfigure
\renewenvironment{figure}[1][2] {
    \expandafter\origfigure\expandafter[H]
} {
    \endorigfigure
}

\usepackage{fixltx2e} % provides \textsubscript
\usepackage[
  backend=biber,
]{biblatex}
\bibliography{paper.bib}

\let\textttOrig=\texttt
\def\texttt#1{\expandafter\textttOrig{\seqsplit{#1}}}
\renewcommand{\seqinsert}{\ifmmode
  \allowbreak
  \else\penalty6000\hspace{0pt plus 0.02em}\fi}

\makeatletter
\let\href@Orig=\href
\def\href@Urllike#1#2{\href@Orig{#1}{\begingroup
    \def\Url@String{#2}\Url@FormatString
    \endgroup}}
\def\href@Notdoi#1#2{\def\tempa{#1}\def\tempb{#2}%
  \ifx\tempa\tempb\relax\href@Urllike{#1}{#2}\else
  \href@Orig{#1}{#2}\fi}
\def\href#1#2{%
  \IfBeginWith{#1}{https://doi.org}%
  {\href@Urllike{#1}{#2}}{\href@Notdoi{#1}{#2}}}
\makeatother

\newlength{\cslhangindent}
\newlength{\csllabelwidth}
\newenvironment{CSLReferences}[3] % #1 hanging-ident, #2 entry spacing
 {% don't indent paragraphs
  \setlength{\parindent}{0pt}
  \ifodd #1 \everypar{\setlength{\hangindent}{\cslhangindent}}\ignorespaces\fi
  \ifnum #2 > 0
  \setlength{\parskip}{#2\baselineskip}
  \fi
 }%
 {}
\usepackage{calc}

\usepackage[top=3.5cm, bottom=3cm, right=1.5cm, left=1.0cm,
            headheight=2.2cm, reversemp, includemp, marginparwidth=4.5cm]{geometry}

\titleformat{\section}
  {\normalfont\sffamily\Large\bfseries}
  {}{0pt}{}
\titleformat{\subsection}
  {\normalfont\sffamily\large\bfseries}
  {}{0pt}{}
\titleformat{\subsubsection}
  {\normalfont\sffamily\bfseries}
  {}{0pt}{}
\titleformat*{\paragraph}
  {\sffamily\normalsize}

\usepackage{fancyhdr}
\makeatletter
\let\ps@plain\ps@fancy
\fancyheadoffset[L]{4.5cm}
\fancyfootoffset[L]{4.5cm}

\definecolor{linky}{rgb}{0.0, 0.5, 1.0}

\newtcolorbox{repobox}
   {colback=red, colframe=red!75!black,
     boxrule=0.5pt, arc=2pt, left=6pt, right=6pt, top=3pt, bottom=3pt}

\newcommand{\ExternalLink}{%
   \tikz[x=1.2ex, y=1.2ex, baseline=-0.05ex]{%
       \begin{scope}[x=1ex, y=1ex]
           \clip (-0.1,-0.1)
               --++ (-0, 1.2)
               --++ (0.6, 0)
               --++ (0, -0.6)
               --++ (0.6, 0)
               --++ (0, -1);
           \path[draw,
               line width = 0.5,
               rounded corners=0.5]
               (0,0) rectangle (1,1);
       \end{scope}
       \path[draw, line width = 0.5] (0.5, 0.5)
           -- (1, 1);
       \path[draw, line width = 0.5] (0.6, 1)
           -- (1, 1) -- (1, 0.6);
       }
   }

\patchcmd{\@maketitle}{center}{flushleft}{}{}
\patchcmd{\@maketitle}{center}{flushleft}{}{}
\patchcmd{\@maketitle}{\LARGE}{\LARGE\sffamily}{}{}
\def\maketitle{{%
  
  \AB@maketitle}}
\makeatletter
\renewcommand\AB@affilsepx{ \protect\Affilfont}
\renewcommand\AB@affilnote[1]{{\bfseries #1}\hspace{3pt}}
\renewcommand{\affil}[2][]%
   {\newaffiltrue\let\AB@blk@and\AB@pand
      \if\relax#1\relax\def\AB@note{\AB@thenote}\else\def\AB@note{#1}%
        \setcounter{Maxaffil}{0}\fi
        \begingroup
        \let\href=\href@Orig
        \let\texttt=\textttOrig
        \let\protect\@unexpandable@protect
        \def\thanks{\protect\thanks}\def\footnote{\protect\footnote}%
        \@temptokena=\expandafter{\AB@authors}%
        {\def\\{\protect\\\protect\Affilfont}\xdef\AB@temp{#2}}%
         \xdef\AB@authors{\the\@temptokena\AB@las\AB@au@str
         \protect\\[\affilsep]\protect\Affilfont\AB@temp}%
         \gdef\AB@las{}\gdef\AB@au@str{}%
        {\def\\{, \ignorespaces}\xdef\AB@temp{#2}}%
        \@temptokena=\expandafter{\AB@affillist}%
        \xdef\AB@affillist{\the\@temptokena \AB@affilsep
          \AB@affilnote{\AB@note}\protect\Affilfont\AB@temp}%
      \endgroup
       \let\AB@affilsep\AB@affilsepx
}
\makeatother

\renewcommand\Affilfont{\sffamily\small\mdseries}
\ifnum 0\ifxetex 1\fi\ifluatex 1\fi=0 % if pdftex
  \usepackage[T1]{fontenc}
  \usepackage[utf8]{inputenc}

\else % if luatex or xelatex
  \ifxetex
    \usepackage{mathspec}
  \else
    \usepackage{fontspec}
  \fi
  \defaultfontfeatures{Ligatures=TeX,Scale=MatchLowercase}

\fi
\IfFileExists{upquote.sty}{\usepackage{upquote}}{}
\IfFileExists{microtype.sty}{%
\usepackage{microtype}
\UseMicrotypeSet[protrusion]{basicmath} % disable protrusion for tt fonts
}{}

\usepackage{hyperref}
\hypersetup{unicode=true,
            pdftitle={What and How of Machine Learning Transparency: Building Bespoke Explainability Tools with Interoperable Algorithmic Components},
            pdfborder={0 0 0},
            breaklinks=true}
\let\addcontentslineOrig=\addcontentsline
\def\addcontentsline#1#2#3{\bgroup
  \let\texttt=\textttOrig\addcontentslineOrig{#1}{#2}{#3}\egroup}
\let\markbothOrig\markboth
\def\markboth#1#2{\bgroup
  \let\texttt=\textttOrig\markbothOrig{#1}{#2}\egroup}
\let\markrightOrig\markright
\def\markright#1{\bgroup
  \let\texttt=\textttOrig\markrightOrig{#1}\egroup}

\usepackage{graphicx,grffile}
\makeatletter
\def\maxwidth{\ifdim\Gin@nat@width>\linewidth\linewidth\else\Gin@nat@width\fi}
\def\maxheight{\ifdim\Gin@nat@height>\textheight\textheight\else\Gin@nat@height\fi}
\makeatother
\setkeys{Gin}{width=\maxwidth,height=\maxheight,keepaspectratio}
\IfFileExists{parskip.sty}{%
\usepackage{parskip}
}{% else
\setlength{\parindent}{0pt}
\setlength{\parskip}{6pt plus 2pt minus 1pt}
}
\providecommand{\tightlist}{%
  \setlength{\itemsep}{0pt}\setlength{\parskip}{0pt}}
\ifx\paragraph\undefined\else
\let\oldparagraph\paragraph
\renewcommand{\paragraph}[1]{\oldparagraph{#1}\mbox{}}
\fi
\ifx\subparagraph\undefined\else
\let\oldsubparagraph\subparagraph
\renewcommand{\subparagraph}[1]{\oldsubparagraph{#1}\mbox{}}
\fi

\title{What and How of Machine Learning Transparency: Building Bespoke
Explainability Tools with Interoperable Algorithmic Components}

        \author[1, 2]{Kacper Sokol}
          \author[1]{Alexander Hepburn}
          \author[1]{Raul Santos-Rodriguez}
          \author[1]{Peter Flach}
    
      \affil[1]{Intelligent Systems Laboratory, University of Bristol,
United Kingdom}
      \affil[2]{ARC Centre of Excellence for Automated Decision-Making
and Society, RMIT University, Australia}
  \date{\vspace{-5ex}}

\begin{document}
\maketitle

\marginpar{
  %\hrule
  \sffamily\small

%  {\bfseries DOI:} \href{https://doi.org/DOI unavailable}{\color{linky}{DOI unavailable}}

%  \vspace{2mm}

  {\bfseries Preprint}
  \vspace{2mm}

  {\bfseries Resources}% Software
  \begin{itemize}
    \setlength\itemsep{0em}
    \item \href{https://github.com/openjournals/jose-reviews/issues/175}{\color{linky}{Review}} \ExternalLink
    \item \href{https://github.com/fat-forensics/Surrogates-Tutorial}{\color{linky}{Repository}} \ExternalLink
%    \item \href{DOI unavailable}{\color{linky}{Archive}} \ExternalLink
  \end{itemize}

  \vspace{2mm}
  {\bfseries License}\\
  Authors of papers retain copyright and release the work under a Creative Commons Attribution 4.0 International License (\href{http://creativecommons.org/licenses/by/4.0/}{\color{linky}{CC BY 4.0}}).
}

\hypertarget{summary}{%
\section{Summary}\label{summary}}

Explainability techniques for data-driven predictive models based on
artificial intelligence and machine learning algorithms allow us to
better understand the operation of such systems and help to hold them
accountable (Sokol \& Flach, 2021a). New transparency approaches are
developed at breakneck speed, enabling us to peek inside these black
boxes and interpret their decisions. Many of these techniques are
introduced as monolithic tools, giving the impression of
one-size-fits-all and end-to-end algorithms with limited
customisability. Nevertheless, such approaches are often composed of
multiple interchangeable modules that need to be tuned to the problem at
hand to produce meaningful explanations (Sokol et al., 2019). This paper
introduces a collection of hands-on training materials -- slides, video
recordings and Jupyter Notebooks -- that provide guidance through the
process of building and evaluating bespoke modular surrogate explainers
for tabular data. These resources cover the three core building blocks
of this technique: interpretable representation composition, data
sampling and explanation generation (Sokol et al., 2019).

\hypertarget{modular-surrogate-explainers}{%
\section{Modular Surrogate
Explainers}\label{modular-surrogate-explainers}}

The training materials introduce the concept of \emph{modular}
explainability algorithms using the example of surrogate explainers for
tabular data. This separation of functionally independent building
blocks allows us to consider the influence of each component, and their
interdependence, on the robustness and faithfulness of the final
explainer. To this end, we review a collection of techniques to evaluate
the quality of the modules and their overall effectiveness. These
metrics can guide the parameterisation of the entire explainability
algorithm, providing an opportunity to tune it to the problem at hand.
All of these insights demonstrate that while surrogate explainers are
model-agnostic and post-hoc -- i.e., they work with any black box and
can be retrofitted into pre-existing predictive models, thus making them
a popular choice for explaining black-box predictions (Ribeiro et al.,
2016) -- using off-the-shelf explainability approaches may result in
subpar performance for individual use cases (Rudin, 2019). Therefore,
understanding how to build a bespoke surrogate explainer that is
suitable for a particular situation is a prerequisite for trustworthy
and meaningful explainability of data-driven systems and their
decisions.

Prior to diving into the practicalities of composing surrogate
explainers, the training materials introduce the concept of algorithmic
explainability of predictive models and discuss the fundamental ideas
behind surrogates for text, image and tabular data. This theoretical
overview is followed by a brief presentation of the software used for
the hands-on modules; \texttt{FAT\ Forensics}\footnote{https://fat-forensics.org/}
is an open source Python package designed for inspecting selected
fairness, accountability and \emph{transparency} aspects of data (and
their features), \emph{models} and \emph{predictions} (Sokol et al.,
2020, 2022). Having covered the basics, the practical coding resources
focus on the three building blocks of surrogate explainers for tabular
data identified by the bLIMEy -- build LIME yourself (Sokol et al.,
2019) -- meta-algorithm:

\begin{itemize}
\tightlist
\item
  interpretable (data) representation composition;
\item
  data sampling; and
\item
  explanation generation (interpretable feature selection, data sample
  weighting, surrogate model training and explanation extraction).
\end{itemize}

These learning modules review some of the interoperable algorithmic
components available at each step, discuss their pros and cons for a
range of applications, guide through their optimal selection strategies
and propose suitable evaluation criteria -- see Figure 1. In particular,
interpretable representations are built with quartile discretisation and
decision trees (Sokol \& Flach, 2020b); data are generated with Gaussian
and mixup sampling (Sokol et al., 2019; Zhang et al., 2018); and
explanations are extracted from linear and tree-based surrogate models
(Sokol \& Flach, 2020a). Notably, these choices determine the type, role
and quality of the resulting explanations composed for black-box
predictions. Therefore, these hands-on materials illustrate how such
interoperable algorithmic building blocks behave in various scenarios
and demonstrate how to use these components to configure robust
explainers with well-known properties and failure modes based on
first-hand observations and a collection of quantitative evaluation
metrics and validation techniques.

\begin{figure}
\centering
\includegraphics[width=0.65\textwidth,height=\textheight]{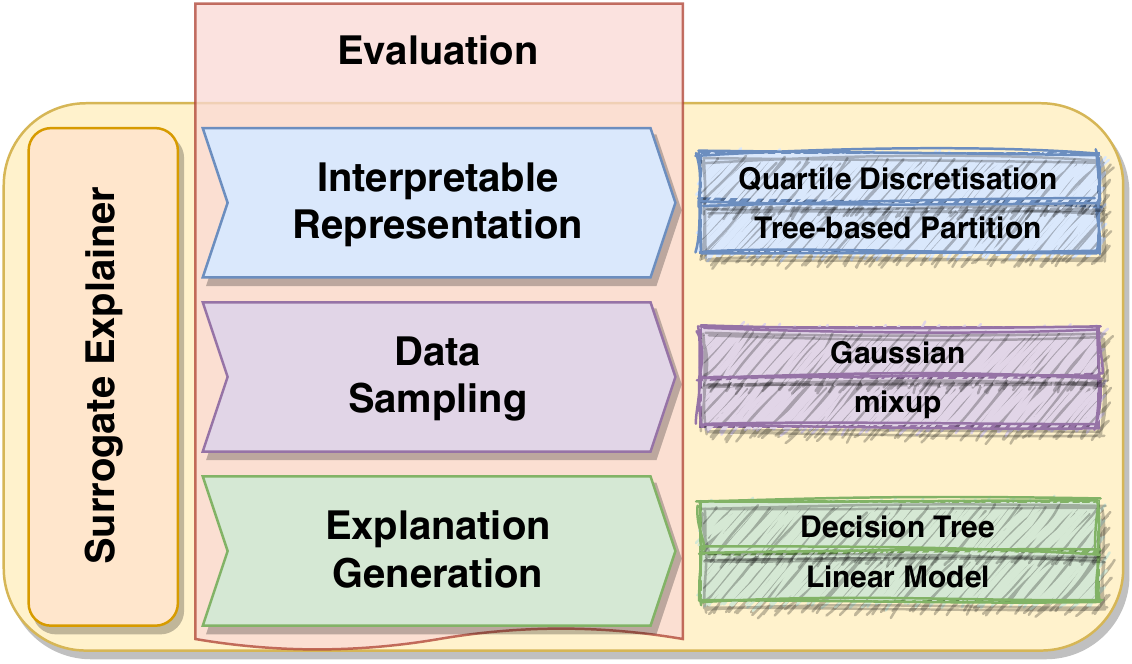}
\caption{Overview of surrogate explainers modularity listing components
specific to tabular data.}
\end{figure}

The introduction to algorithmic explainability; the theoretical overview
of surrogate explainers for text, image and tabular data; and the
outline of \texttt{FAT\ Forensics} are presented in a collection of
\emph{slides} and \emph{instructional video recordings}. The hands-on
materials are delivered with \emph{Jupyter Notebooks} that interweave
textual guidance, code examples and informative plots. All the insights
learnt throughout the practical exercises enable the tutees to create
robust surrogate explainers for an arbitrary black-box predictive model
built for their own tabular data set. The training resources are
designed to appeal and be accessible to an audience with a wide range of
backgrounds and experiences. Active participation in the practical part
requires basic familiarity with Python programming and access to a
computer connected to the Internet, which enables execution of the
Jupyter Notebooks online without installing any software on a personal
machine.

These training materials were used to deliver a hands-on tutorial -- of
the same title -- at the 2020 European Conference on Machine Learning
and Principles and Practice of Knowledge Discovery in Databases
(ECML-PKDD)\footnote{https://events.fat-forensics.org/2020\_ecml-pkdd/},
the recordings of which are available on YouTube\footnote{https://www.youtube.com/playlist?list=PLgdhPOmeUNm0H2XTQECK3wabnDohZURLK}.
Moreover, they inspired a number of interactive sessions at various
summer schools aimed at doctoral students in artificial intelligence and
machine learning, as well as undergraduate lectures, academic
presentations and invited talks. The slides, extra hands-on resources
and video recordings of some of these events are available on the FAT
Forensics Events website\footnote{https://events.fat-forensics.org/}.
The new teaching materials\footnote{https://github.com/fat-forensics/resources/}
additionally cover surrogates for image data -- focusing on the
influence of segmentation granularity and occlusion colour on the
trustworthiness of the resulting explanations (Sokol \& Flach, 2020b) --
and touch upon other explainers such as \emph{permutation importance}
(Breiman, 2001), \emph{individual conditional expectation} (Goldstein et
al., 2015) and \emph{partial dependence} (Friedman, 2001).

Notably, these sessions propelled the improvement, evolution and
expansion of the training resources. In particular, the
programming-focused exercises have been complemented by \emph{no-code}
Jupyter Notebooks that enable interactive experimentation with the
explainers through intuitive \emph{Jupyter Widgets}, thus allowing to
better engage with the audience in a limited time. The same strategy has
been employed for the slides -- by embedding interactive examples based
on widgets -- to which end they have been built with RISE\footnote{https://rise.readthedocs.io/}.
From our experience, the teaching became much more effective when the
ubiquitous PDF slides and Jupyter Notebook programming exercises were
replaced with and/or enriched by formats supporting seamless interaction
with the taught material (in our case achieved through widgets). This
exploration of alternative technologies for building training resources
has also inspired a prototype of a new publishing workflow, where
multiple artefacts such as online documents, slides and computational
notebooks can be composed from a unified collection of source materials
(Sokol \& Flach, 2021b).

\hypertarget{statement-of-need}{%
\section{Statement of Need}\label{statement-of-need}}

The training resources described by this paper introduce a novel
learning paradigm for algorithmic explainability of data-driven
predictive systems based on artificial intelligence and machine learning
techniques. Instead of treating these tools as end-to-end, monolithic
entities whose configuration is only facilitated through the parameters
exposed by their developers, these educational materials look into their
modularity to identify atomic and functionally interoperable building
blocks. By decomposing explainers into their core elements we can better
understand their role and configure them for the application at hand.
Within this purview, such techniques are diagnostic tools that
\emph{only} become explainers when their properties and interpretation
of their outputs are well understood and designed accordingly.
Therefore, to engender trust in data-driven predictive systems, the
employed explainability approaches must be trustworthy themselves in the
first place -- the learning objective underlying the interactive coding
exercises. The training materials achieve these goals by supporting the
following learning outcomes specifically for surrogate explainers (which
were chosen because of their flexibility and popularity):

\begin{itemize}
\tightlist
\item
  identify self-contained algorithmic components of the explainer and
  understand their functions;
\item
  connect these building blocks to the explainability requirements
  unique to the investigated predictive system;
\item
  select appropriate algorithmic components and tune them to the problem
  at hand;
\item
  evaluate these building blocks (in this specific context)
  independently and when joined together to form the final explainer;
  and
\item
  interpret the resulting explanations in view of the uncovered
  properties and limitations of the bespoke explainability algorithm.
\end{itemize}

The modularity and diversity of these training materials -- slides,
video recordings and Jupyter Notebooks -- allow them to be adapted or
directly incorporated into a course on explainable artificial
intelligence and interpretable machine learning, or form the basis of a
range of educational resources such as practical training sessions and
conference tutorials. The module can be taught as is -- reusing the
slides and computational notebooks -- with either bespoke tuition or by
following the prerecorded videos. Alternatively, the narration, figures
and results presented within the notebooks may be shaped into
tailor-made teaching materials. The hands-on resources can also become a
standalone case study supplementing relevant explainability and
interpretability courses. The comprehensive and in-detail presentation
of the topic, covering both the underlying theory and practical aspects,
is suitable for and accessible to undergraduate and postgraduate
students, researchers as well as engineers and data scientists
interested in the subject. This module fills a gap in educational
materials dealing with artificial intelligence and machine learning
transparency by focusing on understanding of the inner workings of these
techniques and the influence of their building blocks on the robustness,
veracity and comprehensibility of explanatory insights into black-box
predictive models.

\hypertarget{acknowledgements}{%
\section{Acknowledgements}\label{acknowledgements}}

This work was partially supported by TAILOR (Trustworthy AI through
Integrating Learning, Optimisation and Reasoning), a project funded by
EU Horizon 2020 research and innovation programme under GA No 952215;
the UKRI Turing AI Fellowship EP/V024817/1; and the ARC Centre of
Excellence for Automated Decision-Making and Society, funded by the
Australian Government through the Australian Research Council (project
number CE200100005).

\hypertarget{references}{%
\section*{References}\label{references}}
\addcontentsline{toc}{section}{References}

\hypertarget{refs}{}
\begin{CSLReferences}{1}{0}
\leavevmode\hypertarget{ref-breiman2001random}{}%
Breiman, L. (2001). Random forests. \emph{Machine Learning},
\emph{45}(1), 5--32. \url{https://doi.org/10.1023/A:1010933404324}

\leavevmode\hypertarget{ref-friedman2001greedy}{}%
Friedman, J. H. (2001). Greedy function approximation: {A} gradient
boosting machine. \emph{Annals of Statistics}, 1189--1232.
\url{https://doi.org/10.1214/aos/1013203451}

\leavevmode\hypertarget{ref-goldstein2015peeking}{}%
Goldstein, A., Kapelner, A., Bleich, J., \& Pitkin, E. (2015). Peeking
inside the black box: {V}isualizing statistical learning with plots of
individual conditional expectation. \emph{Journal of Computational and
Graphical Statistics}, \emph{24}(1), 44--65.
\url{https://doi.org/10.1080/10618600.2014.907095}

\leavevmode\hypertarget{ref-ribeiro2016should}{}%
Ribeiro, M. T., Singh, S., \& Guestrin, C. (2016). {``{W}hy should {I}
trust you?''}: {E}xplaining the predictions of any classifier.
\emph{Proceedings of the 22nd {ACM} {SIGKDD} International Conference on
Knowledge Discovery and Data Mining}, 1135--1144.
\url{https://doi.org/10.1145/2939672.2939778}

\leavevmode\hypertarget{ref-rudin2019stop}{}%
Rudin, C. (2019). Stop explaining black box machine learning models for
high stakes decisions and use interpretable models instead. \emph{Nature
Machine Intelligence}, \emph{1}(5), 206--215.
\url{https://doi.org/10.1038/s42256-019-0048-x}

\leavevmode\hypertarget{ref-sokol2020limetree}{}%
Sokol, K., \& Flach, P. (2020a). {LIMEtree}: {I}nteractively
customisable explanations based on local surrogate multi-output
regression trees. \emph{{arXiv} Preprint {arXiv}:2005.01427}.
\url{https://doi.org/10.48550/arXiv.2005.01427}

\leavevmode\hypertarget{ref-sokol2020towards}{}%
Sokol, K., \& Flach, P. (2020b). Towards faithful and meaningful
interpretable representations. \emph{{arXiv} Preprint
{arXiv}:2008.07007}. \url{https://doi.org/10.48550/arXiv.2008.07007}

\leavevmode\hypertarget{ref-sokol2021explainability}{}%
Sokol, K., \& Flach, P. (2021a). Explainability is in the mind of the
beholder: {E}stablishing the foundations of explainable artificial
intelligence. \emph{{arXiv} Preprint {arXiv}:2112.14466}.
\url{https://doi.org/10.48550/arXiv.2112.14466}

\leavevmode\hypertarget{ref-sokol2021you}{}%
Sokol, K., \& Flach, P. (2021b). You only write thrice: {C}reating
documents, computational notebooks and presentations from a single
source. \emph{Beyond Static Papers: {R}ethinking How We Share Scientific
Understanding in Machine Learning -- {ICLR} 2021 Workshop}.
\url{https://doi.org/10.48550/arXiv.2107.06639}

\leavevmode\hypertarget{ref-sokol2020fatf}{}%
Sokol, K., Hepburn, A., Poyiadzi, R., Clifford, M., Santos-Rodriguez,
R., \& Flach, P. (2020). {FAT Forensics}: {A} {P}ython toolbox for
implementing and deploying fairness, accountability and transparency
algorithms in predictive systems. \emph{Journal of Open Source
Software}, \emph{5}(49), 1904. \url{https://doi.org/10.21105/joss.01904}

\leavevmode\hypertarget{ref-sokol2019blimey}{}%
Sokol, K., Hepburn, A., Santos-Rodriguez, R., \& Flach, P. (2019).
{bLIMEy}: {S}urrogate prediction explanations beyond {LIME}.
\emph{Workshop on Human-Centric Machine Learning ({HCML} 2019) at the
33rd Conference on Neural Information Processing Systems ({NeurIPS})}.
\url{https://doi.org/10.48550/arXiv.1910.13016}

\leavevmode\hypertarget{ref-sokol2022fatf}{}%
Sokol, K., Santos-Rodriguez, R., \& Flach, P. (2022). {FAT Forensics}:
{A} {P}ython toolbox for algorithmic fairness, accountability and
transparency. \emph{Software Impacts}, \emph{14}, 100406.
\url{https://doi.org/10.1016/j.simpa.2022.100406}

\leavevmode\hypertarget{ref-zhang2018mixup}{}%
Zhang, H., Cisse, M., Dauphin, Y. N., \& Lopez-Paz, D. (2018). {mixup}:
{B}eyond empirical risk minimization. \emph{Proceedings of the 6th
International Conference on Learning Representations ({ICLR})}.
\url{https://doi.org/10.48550/arXiv.1710.09412}

\end{CSLReferences}

\end{document}